\documentclass[runningheads]{llncs}

\usepackage[T1]{fontenc}
\usepackage{graphicx}
\usepackage{etoolbox}

\AtBeginDocument{%
  \setlength{\abovedisplayskip}{6pt plus 2pt minus 2pt}
  \setlength{\belowdisplayskip}{6pt plus 2pt minus 2pt}
  \setlength{\abovedisplayshortskip}{3pt plus 1pt minus 1pt}
  \setlength{\belowdisplayshortskip}{3pt plus 1pt minus 1pt}
}

\usepackage{ dsfont } 
\usepackage{amsmath}
\usepackage{caption}
\usepackage{tikz}
\usetikzlibrary{shapes.geometric, arrows.meta, positioning, fit, calc, decorations.pathreplacing}
\usepackage{booktabs}
\usepackage{xcolor}
\begin{document}
\title{A Benchmark of State-Space Models vs. Transformers and BiLSTM-based Models for Historical Newspaper OCR}
%
%
\author{Merveilles AGBETI-MESSAN\inst{1}\orcidID{0009-0007-5024-2138} \and
Pierrick TRANOUEZ\inst{1}\orcidID{0000-0002-1962-0782}  \and
Stéphane NICOLAS\inst{1}\orcidID{0000-0003-0575-6731}  \and
Clément CHATELAIN\inst{2}\orcidID{0000-0001-8377-0630}  \and
Thierry PAQUET\inst{1}\orcidID{0000-0002-2044-7542}}

%


 \authorrunning{M. AGBETI-MESSAN et al.}
%
\institute{
LITIS EA4108, University of Rouen Normandy, France \\
\email{\{komlan-epe-nsin.agbeti-messan, thierry.paquet, pierrick.tranouez, stephane.nicolas\}@univ-rouen.fr}
\and
LITIS EA4108, INSA of Rouen Normandy, France \\
\email{\{clement.chatelain\}@insa-rouen.fr}
}
\titlerunning{A Benchmark for
Historical Newspaper OCR}
\maketitle              
\begin{abstract}
End-to-end OCR for historical newspapers remains challenging, as models must handle long text sequences, degraded print quality, and complex layouts. While Transformer-based recognizers dominate current research, their quadratic complexity limits efficient paragraph-level transcription and large-scale deployment. We investigate linear-time State-Space Models (SSMs), specifically Mamba, as a scalable alternative to Transformer-based sequence modeling for OCR.

We present to our knowledge, the first OCR architecture based on SSMs, combining a CNN visual encoder with bi-directional and autoregressive Mamba sequence modeling, and conduct a large-scale benchmark comparing SSMs with Transformer- and BiLSTM-based recognizers. Multiple decoding strategies (CTC, autoregressive, and non-autoregressive) are evaluated under identical training conditions alongside strong neural baselines (VAN, DAN, DANIEL) and widely used off-the-shelf OCR engines (PERO-OCR, Tesseract OCR, TrOCR, Gemini). 

Experiments on historical newspapers from the Bibliothèque nationale du Luxembourg, with newly released $>99\%$ verified gold-standard annotations, and cross-dataset tests on Fraktur and Antiqua lines, show that all neural models achieve low error rates ($\sim$2\% CER), making computational efficiency the main differentiator. Mamba-based models maintain competitive accuracy while halving inference time and exhibiting superior memory scaling (1.26× vs 2.30× growth at 1000 chars), reaching 6.07\% CER at the severely degraded paragraph level compared to 5.24\% for DAN, while remaining 2.05$\times$ faster. 

We release code, trained models, and standardized evaluation protocols to enable reproducible research and guide practitioners in large-scale cultural heritage OCR available at https://github.com/MarcoPerson/ssm-ocr-benchmark.

\end{abstract}

\section{Introduction}
\label{sec:introduction}

Cultural heritage institutions worldwide face the monumental task of digitizing millions of historical documents to ensure their preservation and accessibility. Among these materials, historical newspapers represent a particularly challenging category due to their physical degradation, non-standardized typography, complex multi-column layouts, and linguistic diversity \cite{springmann2017ocr,impact2011dataset}. The Bibliothèque nationale du Luxembourg (BnL), like its counterparts such as the Bibliothèque nationale de France (BnF) and the Library of Congress, maintains extensive newspaper collections spanning over a century, written in multiple languages (Luxembourgish, French, German), and exhibiting varying degrees of preservation quality. Accurate automatic transcription of these documents through Optical Character Recognition (OCR) is essential not only for full-text search and digital humanities research but also for reducing the prohibitive cost of manual transcription \cite{muehlberger2019transkribus}.

Recent advances in deep learning have dramatically improved OCR performance on contemporary documents, with Transformer or Attention based architectures achieving near-human accuracy on clean printed text. Models such as TrOCR \cite{li2021trocr}, DAN (Document Attention Network) \cite{coquenet2023dan}, and VAN \cite{coquenet2023van} have established new benchmarks by leveraging self-attention mechanisms to capture long-range dependencies in document images. However, these architectures face practical deployment challenges when processing longer text regions. While autoregressive Transformers can handle paragraphs or full pages, they incur substantial computational costs: the $O(n^2)$ complexity of self-attention results in prohibitive memory requirements due to large key-value caches during autoregressive decoding, and sequential token generation leads to high inference latency \cite{gu2023mamba}. These limitations are particularly problematic for heritage institutions digitizing millions of newspaper pages, where batch processing throughput and memory efficiency are critical operational constraints alongside recognition accuracy.

Concurrently, a new family of sequence models has emerged: State-Space Models (SSMs), exemplified by the Mamba architecture \cite{gu2023mamba}. Unlike Transformers, SSMs achieve $O(n)$ computational complexity through selective state-space mechanisms inspired by control theory, enabling efficient processing of long sequences while maintaining a constant memory footprint during inference. Mamba has demonstrated impressive results in natural language processing and computer vision tasks \cite{zhu2024vim,liu2024vmamba}, positioning itself as a viable alternative to Transformers for sequence modeling. Despite this promise, SSMs remain unexplored in the context of text recognition from document images, and their comparative performance against established Transformer and recurrent neural network baselines has not been systematically evaluated on historical documents.

Furthermore, the OCR literature exhibits a persistent divide between autoregressive (AR) and non-autoregressive (non-AR) decoding paradigms. AR models, such as DAN \cite{coquenet2023dan}, generate characters sequentially by conditioning on previously predicted tokens, enabling rich contextual modeling at the cost of inference speed. Non-autoregressive models trained with Connectionist Temporal Classification (CTC) \cite{graves2006ctc} assume conditional independence between time steps, enabling a single forward pass followed by sequential decoding. This offers faster inference than autoregressive models, but may sacrifice accuracy due to the lack of explicit language modeling. This dichotomy has been explored in isolated studies, but a unified evaluation comparing AR and non-AR variants of SSMs, Transformers, and BiLSTM-based architectures on the same dataset and same task is absent from the literature. Additionally, the role of tokenization strategies, character-level versus subword-level (BPE), has not been rigorously assessed for historical document OCR \cite{constum2025daniel}, where archaic spellings and out-of-vocabulary words are prevalent.

To address these gaps, this paper presents a comprehensive benchmark of State-Space Models versus Transformers and BiLSTM-based models for historical newspaper OCR, evaluated at both line and paragraph granularities on the BnL corpus. Our study makes the following contributions:

\begin{itemize}
    \item \textbf{First application of State-Space Models to document OCR:} We introduce a Mamba-based architecture with three decoding variants (CTC, autoregressive, and non-autoregressive) for text recognition in historical documents, demonstrating that SSMs constitute a viable alternative to Transformers for this task and opening a new research direction in heritage document digitization.

    \item \textbf{Systematic comparison across decoding paradigms:} We evaluate six neural architectures spanning three decoding strategies (CTC, autoregressive, and non-autoregressive cross-entropy) and three model families (SSMs, Transformers, and BiLSTMs), providing the first controlled benchmark of these paradigms on historical documents under identical experimental conditions. We additionally compare them against four widely-used production OCR systems (PERO-OCR \cite{kol2021pero}, Tesseract \cite{TessOverview}, TrOCR \cite{li2021trocr}, Gemini \cite{google2025gemini}).

    \item \textbf{Dual-granularity evaluation:} We assess the models at line and paragraph levels, providing insights into how architectures scale when processing longer text regions, a critical consideration for practical deployment in mass digitization workflows.

    \item \textbf{Comprehensive efficiency and robustness analysis:} Beyond recognition accuracy, we report computational metrics (parameters, memory, inference latency, throughput) and investigate model behavior across varying sequence lengths and document degradation levels, enabling practitioners to make informed architectural choices based on operational constraints.

    \item \textbf{High-quality reproducible benchmark:} We release our standardized train/validation/test splits derived from gold-standard human annotations, with test splits manually corrected to $>99\%$ accuracy, along with trained model checkpoints and evaluation code, establishing a reliable foundation for future research in historical newspaper OCR.
\end{itemize}

The remainder of this paper is organized as follows: Section~2 reviews related work in historical OCR, Transformer, BiLSTM, and SSM architectures, as well as paragraph-level recognition. Section~3 describes our methodology, detailing the architectures evaluated and training protocols. Section~4 presents our experimental setup and the BnL datasets. Section~5 reports results and in-depth analyses. Section~6 concludes with directions for future work.

\section{Related Work}
\label{sec:related-work}

\subsection{Historical Document OCR and Newspaper Challenges}

Historical newspapers present unique challenges: image degradation (blur, bleed-through), complex layouts (multi-column, advertisements), and typographic variation across space and time (e.g., Antiqua vs Fraktur typefaces). These factors make recognition accuracy sensitive to preprocessing, segmentation, and the chosen decoding paradigm.

Large-scale digitization often relies on OCR systems mixing layout analysis, line extraction, and line-level recognisers. Transkribus has operationalised trainable handwriting recognition in cultural heritage \cite{muehlberger2019transkribus}. Open-source recognisers like PERO-OCR \cite{kol2021pero} illustrate the importance of robust CTC-style line recognisers in production. Evaluations show modern neural approaches significantly outperform legacy systems like Tesseract \cite{TessOverview} on degraded documents, but performance remains sensitive to domain-specific fine-tuning \cite{springmann2017ocr}.

Benchmarking has been shaped by datasets and competitions: READ-BAD \cite{gruning2017readbad} for baseline detection, DIVA-HisDB \cite{simistira2016diva} for annotated historical pages, and the IMPACT dataset \cite{impact2011dataset} for historical print. ICDAR 2019 included historical document competitions \cite{icdar2019comp}, though evaluations still focus on line-level rather than paragraph-level settings.

\subsection{Sequence Modeling: CTC, Attention, and Transformers}

Most recognizers decompose into a visual encoder and sequence decoder. Three paradigms have emerged for sequence modeling in OCR, differing primarily in how they handle the alignment between visual features and output tokens, and whether they model dependencies between output symbols.

\emph{CTC-based approaches.}
Connectionist Temporal Classification (CTC) \cite{graves2006ctc} sidesteps explicit alignment by marginalizing over all monotonic paths between encoder frames and output symbols. This enables parallel decoding with a conditional independence assumption between output characters. The CRNN \cite{shi2015crnn} combining CNN features with recurrent networks and CTC remains a strong line-level baseline. Recent work explores bidirectional GRUs \cite{bluche2017gru} or convolutional architectures while retaining CTC. VAN \cite{coquenet2023van} demonstrates competitive results on handwritten paragraphs in CTC mode. PyLaia \cite{puigcerver2017pylaia} is a widely-adopted CTC framework for historical documents. While computationally efficient, CTC's independence assumption prevents explicit language modeling, motivating attention-based alternatives.

\emph{Autoregressive attention and Transformers.}
In contrast to CTC's parallel decoding, autoregressive (AR) approaches treat OCR as conditional sequence generation, where each output token is predicted sequentially conditioned on previously generated symbols. This enables rich modeling of linguistic dependencies at the cost of inference latency proportional to sequence length. The Decoupled Attention Network \cite{wang2019dan} decouples alignment from decoding states, SAR \cite{li2019sar} employs attention for visual alignment, and ASTER \cite{shi2018aster} adds rectification for irregular text. Transformer OCR strengthened AR decoding with pre-trained components. TrOCR \cite{li2021trocr} combines pre-trained image and text Transformers. Vision Transformer encoders with Transformer decoders \cite{kim2022donut,bautista2022parseq} demonstrate pure-attention effectiveness. For historical newspapers, quadratic attention complexity and sequential AR decoding make long sequences expensive to process at inference time. DAN \cite{coquenet2023dan} introduced segmentation-free document attention for full-page handwritten recognition.

\emph{Hybrid and non-autoregressive Transformers.}
Seeking to bridge the speed of CTC with the supervision quality of autoregressive models, recent work explores non-autoregressive (NAR) Transformer decoders that predict all positions in parallel using learned queries. Non-autoregressive Transformer decoders predict tokens in parallel. PARSeq \cite{bautista2022parseq} operates in both AR and non-AR modes with competitive accuracy on scene text. DANIEL \cite{constum2025daniel} used BPE (Byte-Pair Encoding) subword tokenization, improving efficiency for morphologically rich languages, and speeding up the decoding process at inference time.

\subsection{Paragraph-Level Recognition}

Paragraph-level recognition aims to recognise text without explicit line segmentation, reducing pipeline brittleness but increasing sequence length and complexity, particularly for historical newspapers where line detection may fail.

\textit{Paragraph models.}
VAN \cite{coquenet2023van} introduces vertical attention to progressively read lines from full paragraph images. The "Scan, Attend and Read" (FPHR) system \cite{bluche2017fphr} uses a CNN encoder with an attention-based recurrent decoder for end-to-end handwritten paragraph recognition, demonstrating competitive results against prior MDLSTM-CTC baselines. SPAN \cite{coquenet2021span} performs paragraph recognition without explicit line break annotations. The ``Start, Follow, Read'' \cite{wigington2018start} learns to read full pages in reading order without requiring line segmentation annotations.

\textit{Challenges for newspapers.}
Paragraph settings amplify issues: longer sequences exacerbate AR decoding latency; CTC must compress long visual streams while maintaining coherence; layout variability increases encoder burden; memory for full-page attention can be prohibitive. Recent architectures handle 100+ character paragraphs \cite{coquenet2023van,constum2025daniel}, though inference speed remains a bottleneck.

\subsection{State-Space Models as Attention Alternatives}

Structured state-space models (SSMs) offer efficient long-range modeling with favorable scaling versus Transformers. S4 \cite{gu2021s4} introduced efficient parameterisation based on HiPPO, showing strong performance on long-range benchmarks. S4D extended this with improved training stability.

\textit{Selective state-space models.}
Mamba \cite{gu2023mamba} introduced input-dependent selective SSMs, enabling content-adaptive propagation while maintaining linear-time processing. Vision Mamba (ViM) \cite{zhu2024vim} and VMamba \cite{liu2024vmamba} adapted SSMs to 2D vision via spatial flattening or directional scanning.

\textit{SSMs for document understanding.}
VL-Mamba \cite{qiao2024vlmamba} explores SSMs for multimodal document tasks. DocMamba \cite{hu2025docmamba} introduces efficient document pre-training, competing with LayoutLM on information extraction at lower cost. For text recognition, some recent works explore SSMs: MambaSTR \cite{zhou2025mambastr} for scene text with masked state space modeling; TextMamba \cite{zhao2025textmamba} for detection; and a TinyML model for Korean handwriting \cite{cui2025context} demonstrating compact Mamba variants for resource-constrained devices.

\textit{Limitations and open challenges in historical document OCR.}
These works focus on scene text or modern handwriting with short sequences (<30 characters) and clean images. No prior work systematically evaluates SSMs on historical newspaper OCR, where sequences are longer (50-500+ characters at paragraph level), images degraded, and typographic variation high. The interaction between SSM backbones and decoding strategies (CTC, AR, non-AR) remains unexplored in OCR.

\textit{Positioning.}
Existing literature provides strong baselines across CTC, attention, and Transformer paradigms, plus paragraph-level models. However, a controlled comparison of modern SSM backbones against Transformers on historical newspapers at both line and paragraph granularities is absent. Our study targets this gap, benchmarking decoding paradigms and model families under a unified protocol on the BnL dataset.

\section{Mamba-based OCR}
\label{sec:methodology}

This section describes our Mamba-based architectures for historical newspaper OCR. We first formalize the sequence recognition problem and three decoding paradigms, then present our model designs with the core Mamba formulation.

\subsection{Preliminaries}
\label{subsec:preliminaries}
We formulate OCR as a sequence prediction task: given an image $\mathbf{I}$ containing text, we predict a character sequence $\mathbf{y} = (y_1, \ldots, y_T)$ with $y_t \in \mathcal{V}$, where $\mathcal{V}$ is the character vocabulary. Three decoding paradigms dominate current approaches, differing in how they factorize $P(\mathbf{y}|\mathbf{I})$. \\

\textbf{Autoregressive (AR) decoding} factorizes $P(\mathbf{y}|\mathbf{I}) = \prod_{t=1}^{T} P(y_t | y_{<t}, \mathbf{I})$, where $y_{<t}$ denotes all previously generated tokens, enabling rich language modeling at the cost of $T$ sequential forward passes.

\textbf{Connectionist Temporal Classification (CTC)} \cite{graves2006ctc} uses a blank symbol to marginalize over monotonic alignments of length $L \geq T$: $P(\mathbf{y}|\mathbf{I}) = \sum_{\mathbf{a} \in \mathcal{A}(\mathbf{y})} \prod_{\ell=1}^{L} P(a_\ell | \mathbf{I})$. CTC assumes conditional independence given the image, allowing efficient parallel frame-wise prediction with greedy or beam search decoding.

\textbf{Non-autoregressive cross-entropy (NAR-CE)} predicts all positions simultaneously using learned position queries: $P(\mathbf{y}|\mathbf{I}) = \prod_{t=1}^{T_{\text{max}}} P(y_t | \mathbf{I})$, where $T_{\text{max}}$ is a fixed maximum length. This combines parallel decoding with direct cross-entropy supervision, avoiding CTC's alignment search.

\subsection{Mamba Formulation}
\label{subsec:mamba}

State Space Models (SSMs) originate from control theory, where they describe dynamic systems through a latent state evolving over time. For an input $x(t) \in \mathds{R}$ and state $h(t) \in \mathds{R}^N$, the continuous dynamics are:
\begin{equation}
h'(t) = \mathbf{A}h(t) + \mathbf{B}x(t), \quad y(t) = \mathbf{C}h(t)
\end{equation}
with $\mathbf{A} \in \mathds{R}^{N \times N}$ (state transition), $\mathbf{B} \in \mathds{R}^{N \times 1}$ (input projection), and $\mathbf{C} \in \mathds{R}^{1 \times N}$ (output mapping). For discrete sequences like text, these are discretized with timescale $\Delta$:
\begin{equation}
h_t = \bar{\mathbf{A}}h_{t-1} + \bar{\mathbf{B}}x_t, \quad y_t = \mathbf{C}h_t
\end{equation}
where $\bar{\mathbf{A}} = \exp(\Delta \mathbf{A})$ and $\bar{\mathbf{B}} = (\Delta \mathbf{A})^{-1}(\exp(\Delta \mathbf{A}) - \mathbf{I}) \cdot \Delta \mathbf{B}$.\\In practice, $\mathbf{A}$, $\mathbf{B}$, $\mathbf{C}$, and $\Delta$ are \textit{learned parameters} optimized during training.

\textbf{Mamba} \cite{gu2023mamba} introduces \textit{selective} SSMs by making $\mathbf{B}$, $\mathbf{C}$, and $\Delta$ input-dependent functions rather than fixed matrices:
\begin{equation}
\mathbf{B} = \text{Linear}_B(\mathbf{x}), \quad \mathbf{C} = \text{Linear}_C(\mathbf{x}), \quad \Delta = \text{softplus}(\text{Linear}_\Delta(\mathbf{x}))
\end{equation}
for input $\mathbf{x} \in \mathds{R}^{L \times D}$. This allows the model to dynamically filter or retain information based on content, analogous to attention mechanisms, while maintaining $O(n)$ linear complexity via hardware-aware parallel scanning. All linear projections ($\text{Linear}_B$, $\text{Linear}_C$, $\text{Linear}_\Delta$) are trainable neural layers that adapt to the OCR task during backpropagation.

\subsection{Mamba-based Architectures}
\begin{figure*}[!ht]
\centering
\resizebox{\textwidth}{!}{
\begin{tikzpicture}[
    node distance=0.7cm,
    procbox/.style={rectangle, draw, rounded corners, minimum width=3cm, minimum height=0.7cm, align=center, fill=#1, font=\sffamily\small, line width=0.8pt},
    procbox/.default={blue!20},
    bigprocbox/.style={rectangle, draw, rounded corners, minimum width=3cm, minimum height=1.2cm, align=center, fill=#1, font=\sffamily\small, line width=0.8pt},
    databox/.style={rectangle, draw, dashed, minimum width=3cm, minimum height=0.6cm, align=center, fill=white, font=\sffamily\small\itshape, line width=0.6pt},
    arrow/.style={->, >=stealth, thick},
    dashedarrow/.style={->, >=stealth, thick, dashed},
    label/.style={font=\sffamily\bfseries\footnotesize}
]
\node[font=\sffamily\small\itshape] (input_label) {Input Image ($H \times W \times 3$)};
\node[inner sep=0pt, below=0.05cm of input_label] (input) {
    \includegraphics[width=8cm]{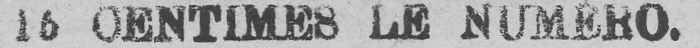}
};
\node[bigprocbox={blue!30}, below=0.5cm of input, minimum width=4cm] (encoder) {
    \textbf{CNN Encoder}\\
    5 layers (Conv+BN+Pool)
};
\node[procbox={blue!20}, below=of encoder, minimum width=4cm] (pos2d) {
    2D Positional Encoding\\
    + Flatten to sequence
};
\node[databox, below=of pos2d, minimum width=4cm] (features_flat) {
    Visual Features\\
    $L \times 256$
};
\node[bigprocbox={purple!30}, below=of features_flat, minimum width=4cm, minimum height=1.3cm] (bimamba) {
    \textbf{Bidirectional Mamba}\\
    $\text{Mamba}_{fwd} + \text{Mamba}_{bwd}$\\
    \scriptsize{+ LayerNorm + FFN}
};
\node[databox, below=of bimamba, minimum width=4cm] (encoded) {
    Encoded Features\\
    $L \times 256$
};
\coordinate (branch) at ($(encoded.south) + (0,-0.8)$);
\node[label, below=0.5cm of branch, xshift=-6.5cm, text=orange!70!black] (ar_label) {\textbf{Mamba-AR Decoder}};
\node[databox, below=0.2cm of ar_label, minimum width=3.5cm] (ar_embed_data) {
    Token Embeddings
};
\node[bigprocbox={orange!30}, below=0.3cm of ar_embed_data, minimum width=3.5cm, minimum height=1.5cm] (ar_mamba) {
    \textbf{4× Mamba Layers}\\
    (unidirectional)\\
    Causal masking
};
\node[procbox={orange!20}, below=0.3cm of ar_mamba, minimum width=3.5cm, minimum height=0.8cm] (ar_proj) {
    Linear Projection\\
    \textbf{Cross-Entropy Loss}
};
\node[procbox={cyan!20}, below=0.3cm of ar_proj, minimum width=3.5cm, minimum height=0.6cm] (ar_softmax) {
    Softmax
};
\node[databox, below=0.2cm of ar_softmax, minimum width=3.5cm] (ar_probs) {
    $|\mathcal{V}|$ probabilities
};
\node[procbox={teal!20}, below=0.2cm of ar_probs, minimum width=3.5cm, minimum height=0.6cm] (ar_argmax) {
    Argmax
};
\node[databox, below=0.2cm of ar_argmax, minimum width=3.5cm, fill=green!5] (ar_text) {
    \textbf{Output Text}\\
    "16 CENTIMES LE NUM..."
};
\node[below=0.05cm of ar_text, font=\sffamily\scriptsize, text width=3.5cm, align=center] (ar_note) {
    \textit{Sequential decoding}
};

\node[below=3cm of branch, xshift=-3.25cm, font=\sffamily\Large\bfseries, text=red!70!black] (or1) {OR};

\node[label, below=0.5cm of branch, text=orange!70!black] (nar_label) {\textbf{Mamba-NAR Decoder}};
\node[databox, below=0.2cm of nar_label, minimum width=3.5cm] (nar_queries) {
    Static Queries\\
    $T_{\max} \times 256$
};
\node[bigprocbox={orange!30}, below=0.3cm of nar_queries, minimum width=3.5cm, minimum height=1.5cm] (nar_mamba) {
    \textbf{4× Mamba Layers}\\
    (unidirectional)\\
    Parallel prediction
};
\node[procbox={orange!20}, below=0.3cm of nar_mamba, minimum width=3.5cm, minimum height=0.8cm] (nar_proj) {
    Linear Projection\\
    \textbf{Masked CE Loss}
};
\node[procbox={cyan!20}, below=0.3cm of nar_proj, minimum width=3.5cm, minimum height=0.6cm] (nar_softmax) {
    Softmax
};
\node[databox, below=0.2cm of nar_softmax, minimum width=3.5cm] (nar_probs) {
    $T_{\max} \times |\mathcal{V}|$ probabilities
};
\node[procbox={teal!20}, below=0.2cm of nar_probs, minimum width=3.5cm, minimum height=0.6cm] (nar_argmax) {
    Argmax
};
\node[databox, below=0.2cm of nar_argmax, minimum width=3.5cm, fill=green!5] (nar_text) {
    \textbf{Output Text}\\
    "16 CENTIMES LE NUMÉRO."
};
\node[below=0.05cm of nar_text, font=\sffamily\scriptsize, text width=3.5cm, align=center] (nar_note) {
    \textit{Single parallel pass}
};

\node[below=3cm of branch, xshift=3.25cm, font=\sffamily\Large\bfseries, text=red!70!black] (or2) {OR};

\node[label, below=0.5cm of branch, xshift=6.5cm, text=orange!70!black] (ctc_label) {\textbf{Mamba-CTC Decoder}};
\node[databox, below=0.2cm of ctc_label, minimum width=3.5cm] (ctc_note_top) {
    Vocab: $\mathcal{V} \cup \{\text{blank}\}$
};
\node[bigprocbox={orange!30}, below=0.3cm of ctc_note_top, minimum width=3.5cm, minimum height=1.5cm] (ctc_mamba) {
    \textbf{4× Mamba Layers}\\
    (unidirectional)\\
    Frame-level prediction
};
\node[procbox={orange!20}, below=0.3cm of ctc_mamba, minimum width=3.5cm, minimum height=0.8cm] (ctc_proj) {
    Linear Projection\\
    \textbf{CTC Loss}
};
\node[procbox={cyan!20}, below=0.3cm of ctc_proj, minimum width=3.5cm, minimum height=0.6cm] (ctc_softmax) {
    Softmax
};
\node[databox, below=0.2cm of ctc_softmax, minimum width=3.5cm] (ctc_probs) {
    $L \times |\mathcal{V}'|$ probabilities
};
\node[procbox={teal!20}, below=0.2cm of ctc_probs, minimum width=3.5cm, minimum height=0.6cm] (ctc_decode_box) {
    CTC Decoding + Argmax
};
\node[databox, below=0.2cm of ctc_decode_box, minimum width=3.5cm, fill=green!5] (ctc_text) {
    \textbf{Output Text}\\
    "16 CENTIMES LE NUMÉRO."
};
\node[below=0.05cm of ctc_text, font=\sffamily\scriptsize, text width=3.5cm, align=center] (ctc_note) {
    \textit{Alignment collapse}
};
\draw[arrow] (input.south) -- (encoder);
\draw[arrow] (encoder) -- (pos2d);
\draw[arrow] (pos2d) -- (features_flat);
\draw[arrow] (features_flat) -- (bimamba);
\draw[arrow] (bimamba) -- (encoded);
\draw[dashedarrow] (encoded.south) -- (branch);
\draw[arrow] (branch) -| (ar_label.north);
\draw[arrow] (branch) -- (nar_label.north);
\draw[arrow] (branch) -| (ctc_label.north);
\draw[arrow] (ar_label) -- (ar_embed_data);
\draw[arrow] (ar_embed_data) -- (ar_mamba);
\draw[arrow] (ar_mamba) -- (ar_proj);
\draw[arrow] (ar_proj) -- (ar_softmax);
\draw[arrow] (ar_softmax) -- (ar_probs);
\draw[arrow] (ar_probs) -- (ar_argmax);
\draw[arrow] (ar_argmax) -- (ar_text);
\draw[arrow] (nar_label) -- (nar_queries);
\draw[arrow] (nar_queries) -- (nar_mamba);
\draw[arrow] (nar_mamba) -- (nar_proj);
\draw[arrow] (nar_proj) -- (nar_softmax);
\draw[arrow] (nar_softmax) -- (nar_probs);
\draw[arrow] (nar_probs) -- (nar_argmax);
\draw[arrow] (nar_argmax) -- (nar_text);
\draw[arrow] (ctc_label) -- (ctc_note_top);
\draw[arrow] (ctc_note_top) -- (ctc_mamba);
\draw[arrow] (ctc_mamba) -- (ctc_proj);
\draw[arrow] (ctc_proj) -- (ctc_softmax);
\draw[arrow] (ctc_softmax) -- (ctc_probs);
\draw[arrow] (ctc_probs) -- (ctc_decode_box);
\draw[arrow] (ctc_decode_box) -- (ctc_text);
\node[below=0.7cm of nar_text, xshift=-6cm] (legend_title) {\textbf{Legend:}};
\node[procbox={blue!30}, right=0.2cm of legend_title, minimum width=2.2cm, minimum height=0.5cm] (legend_enc) {\textcolor{blue!70!black}{Visual Encoder}};
\node[procbox={purple!30}, right=0.2cm of legend_enc, minimum width=2.2cm, minimum height=0.5cm] (legend_bi) {\textcolor{purple!70!black}{BiMamba}};
\node[procbox={orange!30}, right=0.2cm of legend_bi, minimum width=2.2cm, minimum height=0.5cm] (legend_dec) {\textcolor{orange!70!black}{Decoders}};
\node[databox, right=0.2cm of legend_dec, minimum width=1.5cm, minimum height=0.5cm] (legend_data) {Data};
\end{tikzpicture}
}
\caption{\textbf{Mamba-based OCR architecture.} \textcolor{blue!70!black}{Shared visual encoder} (blue) extracts features, \textcolor{purple!70!black}{bidirectional Mamba connector} (purple) models global context, and three \textcolor{orange!70!black}{decoder variants} (orange) implement different decoding strategies: autoregressive (AR), non-autoregressive with static queries (NAR), and CTC alignment.}
\label{fig:mamba-architectures}
\end{figure*}
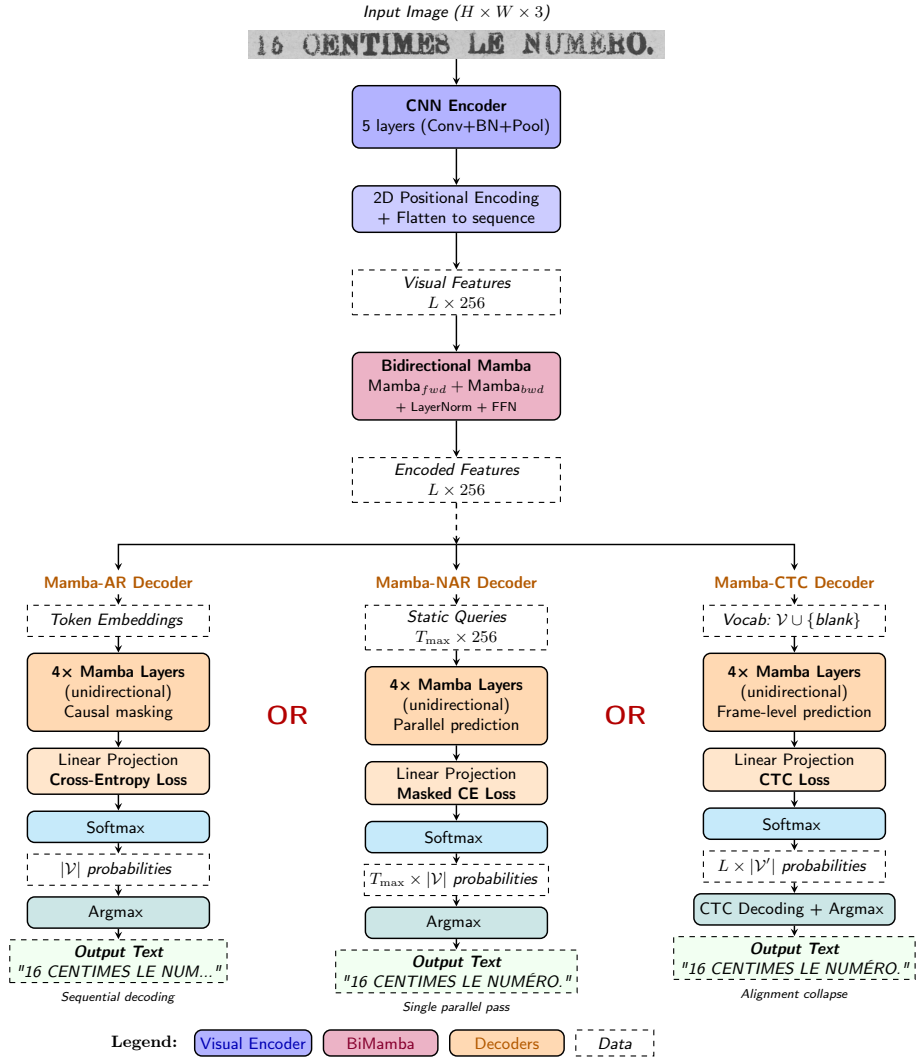

\vspace{-0.5mm}
\label{subsec:mamba-architectures}
We propose three Mamba-based models sharing a common visual encoder and connector but employing different decoding strategies. Figure~\ref{fig:mamba-architectures} provides an overview of all variants.

\subsubsection{\textcolor{blue!70!black}{Shared CNN Encoder}}
\label{subsubsec:encoder}
To ensure fair comparison, we adopt the CNN encoder from DAN \cite{coquenet2023dan}: five convolutional layers with batch normalization and max-pooling produce a 2D feature map of dimension $H' \times W' \times 256$, flattened to a sequence $\mathbf{x} \in \mathds{R}^{L \times 256}$ where $L = H' \times W'$.

\subsubsection{\textcolor{purple!70!black}{Bidirectional Mamba Context Modeling}}
\label{subsubsec:bidirectional}
A single bidirectional Mamba connector processes the visual sequence to capture global context in both directions. This layer employs a residual architecture with normalization and feedforward components. Given the input sequence $\mathbf{x} \in \mathds{R}^{L \times 256}$, we first apply layer normalization and a GELU-activated linear transformation:
\begin{equation}
\tilde{\mathbf{x}} = \text{GELU}(\text{Linear}_1(\text{LayerNorm}_1(\mathbf{x})))
\end{equation}
The transformed sequence is then processed bidirectionally using the same Mamba block. Forward processing propagates information left-to-right, while backward processing reverses the sequence, applies Mamba, and reverses the output:
\begin{equation}
\mathbf{h}^{\text{fwd}} = \text{Mamba}(\tilde{\mathbf{x}}), \quad \mathbf{h}^{\text{bwd}} = \text{flip}(\text{Mamba}(\text{flip}(\tilde{\mathbf{x}})))
\end{equation}
where $\text{flip}(\cdot)$ reverses the temporal dimension. The forward and backward outputs are combined via addition (not concatenation), normalized, and passed through a feedforward layer with residual connections.

\vspace{-0.35cm}

\subsubsection{\textcolor{orange!70!black}{Decoding Variants}}
\label{subsubsec:decoding}
All variants use four unidirectional Mamba layers with an expansion factor of 6 (projecting the 256-dimensional input to an inner dimension of 1536) and a state dimension of 256 (the recurrent state size) before their respective output heads.

\emph{\textcolor{orange!70!black}{Mamba-AR.}} 
At each step $t$, visual context $\mathbf{h} \in \mathds{R}^{L \times 256}$ is concatenated with token embeddings $[\mathbf{e}_1, \ldots, \mathbf{e}_t] \in \mathds{R}^{t \times 256}$ along the sequence dimension. The 4 Mamba layers process this with causal masking, fusing modalities at input level rather than via explicit cross-attention. Training uses teacher forcing with cross-entropy; inference is sequential.
\vspace{-0.2cm}
\begin{equation}
\mathbf{d}_t = \text{Mamba}_{\text{dec}}([\mathbf{h}; \mathbf{e}_{1:t}]), \quad P(y_t | y_{<t}) = \text{softmax}(\mathbf{W}_{\text{AR}} \mathbf{d}_t)
\end{equation}

\emph{\textcolor{orange!70!black}{Mamba-CTC.}} 
Projects Mamba outputs to vocabulary $|\mathcal{V}'|$ with CTC loss and greedy decoding.
\begin{equation}
\mathbf{z} = \text{Mamba}_{\text{dec}}(\mathbf{h}), \quad P(a_\ell | \mathbf{I}) = \text{softmax}(\mathbf{W}_{\text{CTC}} \mathbf{z}_\ell)
\end{equation}

\emph{\textcolor{orange!70!black}{Mamba-NAR.}}
Uses $T_{\max}=500$ learned queries $\mathbf{Q} \in \mathds{R}^{T_{\max} \times 256}$ for parallel prediction with cross-entropy. The decoder emits $T_{\max}$ logits in a single pass.
\begin{equation}
\mathbf{d} = \text{Mamba}_{\text{dec}}([\mathbf{h}; \mathbf{Q}]), \quad P(y_t | \mathbf{I}) = \text{softmax}(\mathbf{W}_{\text{NAR}} \mathbf{d}_t)
\end{equation}

Table~\ref{tab:model-comparison} summarizes all models with parameter counts.

\begin{table*}[t]
\centering
\caption{Comparison of evaluated architectures. Mamba variants differ only in decoding strategy.}
\label{tab:model-comparison}
\resizebox{\textwidth}{!}{
\begin{tabular}{@{}llllcc@{}}
\toprule
\textbf{Model} & \textbf{Backbone} & \textbf{Decoding} & \textbf{Tokenization} & \textbf{Params (M)} & \textbf{Pre-trained?} \\
\midrule
\multicolumn{6}{l}{\textit{Mamba-based (ours)}} \\
Mamba-CTC & CNN + SSM & CTC & Character & $\sim$14.1 & No \\
Mamba-AR & CNN + SSM & Autoregressive & Character & $\sim$14.1 & No \\
Mamba-NAR & CNN + SSM & NAR (CE) & Character & $\sim$14.1 & No \\
\midrule
\multicolumn{6}{l}{\textit{Transformer baselines}} \\
VAN \cite{coquenet2023van} & CNN + Transformer + LSTM & CTC & Character & $\sim$2.4 & No \\
DAN \cite{coquenet2023dan} & CNN + Transformer & Autoregressive & Character & $\sim$7.1 & No \\
DANIEL \cite{constum2025daniel} & CNN + Transformer (BART) & Autoregressive & BPE (XLM-R) & $\sim$146.5 & Partially \\
\midrule
\multicolumn{6}{l}{\textit{Off-the-shelf systems}} \\
PERO-OCR \cite{kol2021pero} & CNN + BiLSTM & CTC & Character & -- & Yes (EU newspapers) \\
Tesseract v5 \cite{TessOverview} & LSTM & -- & -- & -- & Yes (generic) \\
TrOCR (zero-shot) \cite{li2021trocr} & ViT + Transformer & Autoregressive & BPE & $\sim$334 & Yes (LAION, etc.) \\
TrOCR (fine-tuned) \cite{li2021trocr} & ViT + Transformer & Autoregressive & BPE & $\sim$334 & Yes + BnL fine-tuning \\
Gemini \cite{google2025gemini} & VLM (proprietary) & Prompted & -- & -- & Yes (proprietary) \\
\bottomrule
\end{tabular}
}
\end{table*}

\section{Experimental Setup}
\label{sec:experimental-setup}
\vspace{-0.1cm}
\subsection{Datasets}
\label{subsec:datasets}

\subsubsection{BnL Historical Newspaper Corpus}
\label{subsubsec:bnl-corpus}

We evaluate on two datasets (one at line-level, the other at paragraph-level) from the Bibliothèque nationale du Luxembourg's public domain historical newspaper collection\footnote{\url{https://data.bnl.lu/data/historical-newspapers/}}, covering pre-1878 publications in German, French, and Luxembourgish.

\textit{Line-Level Dataset.}
The line-level dataset contains 33,000 manually transcribed text lines with double-keying verification ($>$99.95\% accuracy), split by font type: 14,000 Fraktur and 19,000 Antiqua. We use the following train/validation/test splits: 23,314/1,665/8,328 lines. The character vocabulary size is 129, including 19th-century specific glyphs (accents, ligatures, punctuation).

\begin{table}[ht!]
\small
\centering
\begin{minipage}{0.48\textwidth}
\centering
\caption{Line-level dataset statistics.}
\label{tab:line-stats}
\resizebox{\textwidth}{!}{
\begin{tabular}{@{}lccc@{}}
\toprule
\textbf{Split} & \textbf{Fraktur} & \textbf{Antiqua} & \textbf{Total} \\
\midrule
Train & 9,817 & 13,497 & 23,314 \\
Validation & 701 & 964 & 1,665 \\
Test & 3,507 & 4,821 & 8,328 \\
\bottomrule
\end{tabular}
}
\end{minipage}%
\hfill
\begin{minipage}{0.48\textwidth}
\centering
\caption{Paragraph-level dataset statistics (1--135 lines/paragraph).}
\label{tab:para-stats-full}
\resizebox{\textwidth}{!}{
\begin{tabular}{@{}lccc@{}}
\toprule
\textbf{Metric} & \textbf{Train} & \textbf{Valid} & \textbf{Test} \\
\midrule
Total Paragraphs & 20,604 & 1,795 & 1,540 \\
Total Lines & 308,692 & 26,844 & 22,668 \\
Avg Lines/Para & 14.98 & 14.95 & 14.72 \\
Avg Chars/Para & 617.86 & 616.59 & 605.68 \\
Avg Chars/Line & 40.31 & 40.30 & 40.22 \\
\bottomrule
\end{tabular}
}
\end{minipage}
\end{table}

\textit{Paragraph-Level Dataset.}
Derived from \textit{L'Indépendance Luxembourgeoise} (1877), this dataset comprises 23,939 text regions extracted from 304 newspaper issues. Table~\ref{tab:para-stats-full} provides full corpus statistics.

For our main experiments, we filter to severely degraded paragraphs with 1--10 lines, which covers 86\% of all paragraphs while keeping sequence length under 1000 characters. This yields 13,013 training, 1,156 validation, and 976 test paragraphs. The character vocabulary for paragraph data is 121. We additionally train DANIEL on the full range (1--135 lines) to assess extreme-length handling.

\textit{Ground Truth Creation.}
While METS/ALTO files provide layout coordinates, their embedded transcriptions are unreliable. For training/validation sets, we generate pseudo-ground truth using ROVER \cite{fiscus1997rover} to combine outputs from Gemini, PERO-OCR, and Tesseract with weighted voting (Gemini: 5, PERO: 3, Tesseract: 3). This weighting reflects Gemini's superior accuracy while allowing PERO and Tesseract to override it only when they agree. We acknowledge potential circularity from evaluating Gemini on ROVER-derived labels; however, our manually corrected gold-standard test set (976 paragraphs) eliminates this bias for final metrics.

\textit{Image Preprocessing.}
All images are normalized to 300 DPI. Line and paragraph images are padded to minima of 100 pixels (height) and 1000 pixels (width) respectively, while preserving aspect ratio.

\subsection{Baseline Systems}
\label{subsec:baselines}
We compare against strong baselines representing different architectural families and usage scenarios.

\subsubsection{Transformer-based Models} \textbf{ } \\

\textbf{DAN} \cite{coquenet2023dan} is our primary autoregressive Transformer baseline. It shares the same CNN encoder as our Mamba models, followed by an 8-layer Transformer decoder with 4 attention heads, ensuring direct architectural comparability. \textbf{DANIEL} \cite{constum2025daniel} extends DAN with Byte-Pair Encoding (BPE) subword tokenization using XLM-RoBERTa and a 4-layer BART decoder \cite{lewis2020bart}, testing whether subword tokens better handle historical spelling variations than character-level models. \textbf{VAN} \cite{coquenet2023van} is our CTC-based Transformer baseline. It augments the shared CNN encoder with an implicit line segmentation attention mechanism followed by a unidirectional LSTM before CTC projection, demonstrating competitive accuracy on long sequences without autoregressive decoding.

\subsubsection{Off-the-shelf OCR Systems} \textbf{ } \\

\textbf{PERO-OCR} \cite{kol2021pero} is a production-grade CNN-BiLSTM-CTC system pre-trained on European historical newspapers. We evaluate it zero-shot using the \texttt{pero\_eu\_cz\_print\_newspapers\_2022-09-26} checkpoint, representing out-of-the-box performance. \textbf{Tesseract v5} \cite{TessOverview} provides a lower-bound baseline, illustrating the gap between classical image processing with LSTM-based recognition and modern neural approaches, on degraded historical documents. \textbf{TrOCR} \cite{li2021trocr} combines a Vision Transformer encoder with a Transformer decoder, pre-trained on large-scale document data. We evaluate two configurations: zero-shot inference and fine-tuning on our BnL training set. \textbf{Gemini} \cite{google2025gemini} represents the frontier of foundation model OCR. We query the Gemini API with structured prompts requesting line-by-line text extraction in XML format. Given its proprietary nature, these results indicate current VLM capabilities rather than providing a reproducible benchmark.

\subsection{Training Configuration}
\label{subsec:training-config}

\emph{Pre-training Strategy.}
To mitigate overfitting on limited BnL data, we pre-train DAN and all Mamba variants on 1 million synthetic lines generated from French Wikipedia articles. For paragraph-level Mamba-AR and DAN, we adopt curriculum learning: pre-training proceeds on synthetic paragraphs of gradually increasing length (1--10 lines). Even after transitioning to real BnL data, we maintain 10\% synthetic paragraphs in each batch to preserve domain robustness. DANIEL and TrOCR leverage their existing pre-training, while VAN requires none for line-level CTC \cite{coquenet2023van}. All CTC models (Mamba-CTC, VAN, PERO-OCR) use greedy decoding without external language models. Beam search with LM fusion could improve accuracy but would complicate comparison across paradigms.

\emph{Optimization and Hyperparameters.}
All models use AdamW with learning rates model-specific: $10^{-4}$ for DAN, VAN, and Mamba variants; $10^{-5}$ for DANIEL (fine-tuning BART); $10^{-5}$ with cosine annealing to $10^{-10}$ for TrOCR (500-step warmup). Batch size is 4 for line-level and 1 for paragraph-level recognition due to memory constraints.

\emph{Data Augmentation.}
We apply augmentations (probability 0.5) to simulate historical document variability: DPI rescaling, perspective distortion, elastic deformation, morphological operations, color jitter, Gaussian blur/noise, and sharpening, all applied in random order.

\subsection{Evaluation Metrics}
\label{subsec:metrics}

\emph{Recognition Quality.}
We report Character Error Rate (CER) and Word Error Rate (WER) as normalized Levenshtein distance:
$\text{CER} = \frac{S + D + I}{N} \times 100\%$, $\text{WER} = \frac{S_w + D_w + I_w}{N_w} \times 100\%$.
CER is our primary metric as it directly measures transcription accuracy independent of tokenization.

\emph{Computational Efficiency.}
We measure inference latency (ms/image, on A100), throughput (images/sec), and peak GPU memory (GB) during inference, reflecting deployment constraints for large-scale digitization.



\section{Results}
\label{sec:results}

\subsection{Line-Level Recognition}
\label{subsec:line-results}

Tables~\ref{tab:line-antiqua} and~\ref{tab:line-fraktur} present line-level recognition results on Antiqua and Fraktur test sets respectively.

\emph{Accuracy.} Neural models trained on BnL data achieve competitive recognition quality on Antiqua (1.83--2.66\% CER), with Mamba-AR and DAN tied for best accuracy. VAN achieves lowest WER (3.39\%). On Fraktur, performance degrades slightly (3.01--3.20\% CER), with VAN and Mamba-NAR achieving best results. DANIEL's substantial drop on Fraktur (6.18\% vs 2.37\% CER on Antiqua) likely reflects BPE tokenizer vocabulary mismatch. Off-the-shelf systems perform poorly: zero-shot TrOCR fails catastrophically (15.46--35.98\% CER), while fine-tuned TrOCR achieves 3.62--4.71\% CER at prohibitive cost. Gemini (7.23--10.50\% CER) lags specialized models despite massive scale.

\begin{figure*}[!t]
\centering
\begin{minipage}{\textwidth}
\centering
\includegraphics[width=\textwidth]{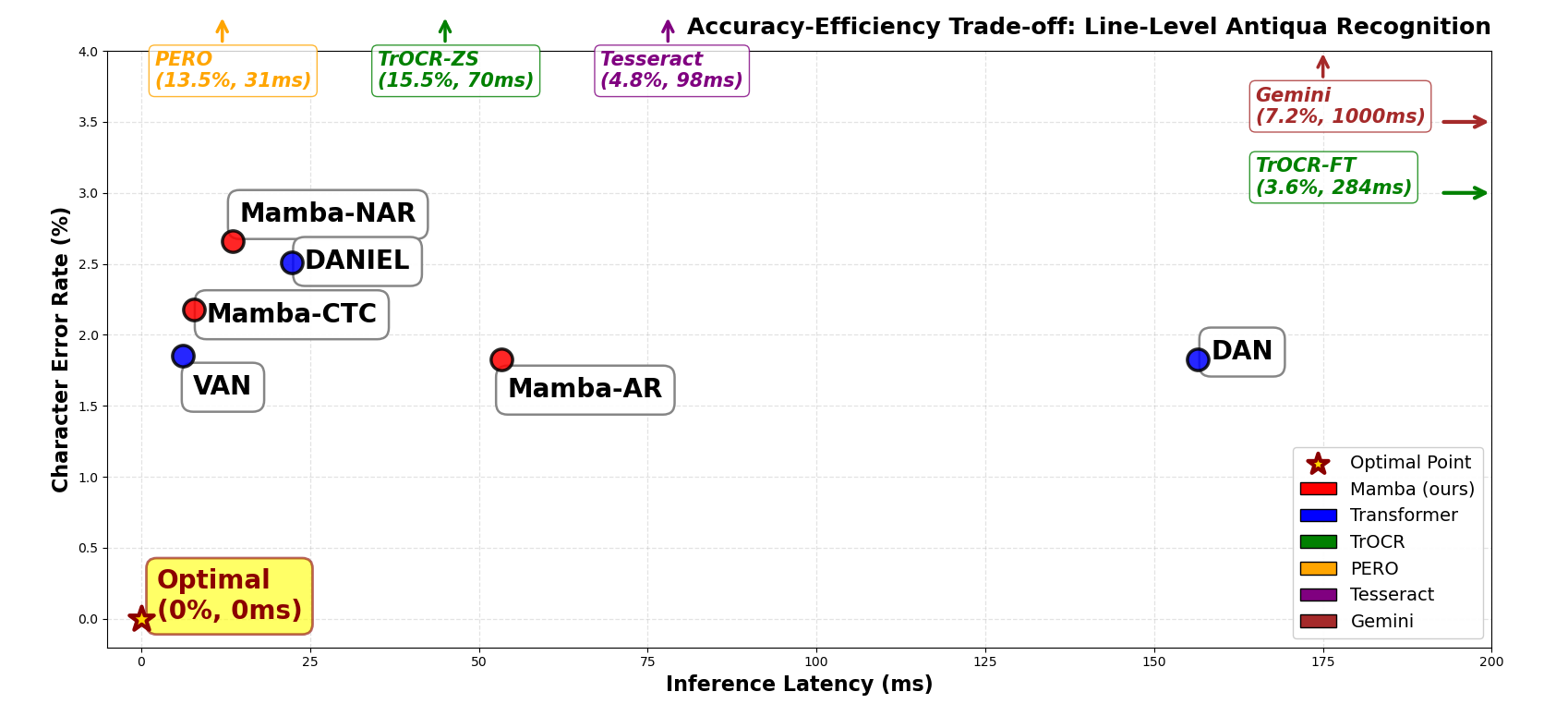}
\end{minipage}%
\hfill
\begin{minipage}{\textwidth}
\centering
\includegraphics[width=\textwidth]{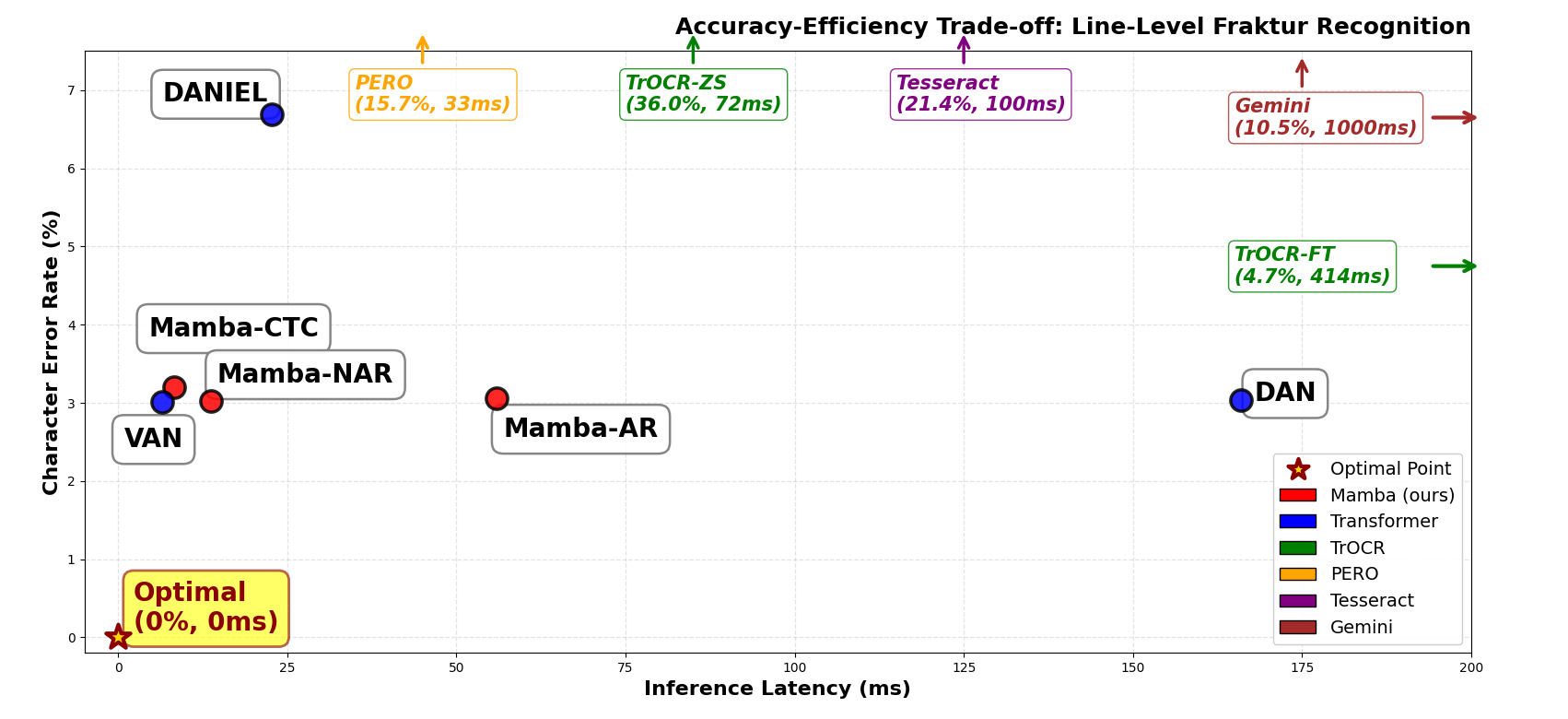}
\end{minipage}
\caption{Accuracy vs latency trade-off on line-level recognition. \textbf{Left:} Antiqua test set, Mamba-AR achieves DAN-level accuracy with 2.9× lower latency. \textbf{Right:} Fraktur test set, VAN and Mamba-NAR provide optimal speed-accuracy balance for Gothic script.}
\label{fig:pareto-line}
\end{figure*}

\emph{Efficiency.} VAN dominates throughput (161.5 img/s on Antiqua) with lowest latency (6.19 ms), followed by Mamba-CTC (129.2 img/s). Autoregressive models incur substantial overhead: DAN requires 156.5 ms per line (25× slower than VAN), while Mamba-AR achieves 53.3 ms (2.9× faster than DAN). Figure~\ref{fig:pareto-line} visualizes the accuracy-latency trade-off for both scripts.

\begin{table*}[ht!]
\centering
\caption{Line-level recognition performance on Antiqua test set 
(Best in \textbf{bold}).}
\small
\label{tab:line-antiqua}
\resizebox{\textwidth}{!}{
\begin{tabular}{@{}lccccccc@{}}
\toprule
\textbf{Model} & \textbf{CER }  & \textbf{WER }  & \textbf{Latency }  & \textbf{Throughput }  & \textbf{Inf Mem } & \textbf{Train Mem } \\
 & \textbf{(\%, $\downarrow$)}  & \textbf{(\%, $\downarrow$)}  & \textbf{(ms, $\downarrow$)} & \textbf{(img/s, $\uparrow$)} & \textbf{(MB)} & \textbf{(GB)} \\
\midrule
\multicolumn{7}{l}{\textit{Mamba-based (ours)}} \\
Mamba-CTC & 2.18 & 5.16 & 7.74 & 129.2 & 543 & 7.17 \\
Mamba-AR & \textbf{1.83} & \textbf{3.84} & 53.3 & 18.7 & 2898 & 8.06 \\
Mamba-NAR & 2.66 & 8.26 & 13.6 & 73.5 & 960 & 6.31 \\
\midrule
\multicolumn{7}{l}{\textit{Transformer baselines}} \\
VAN & 1.85 & \textbf{3.39} & \textbf{6.19} & \textbf{161.5} & 1379 & 14.14 \\
DAN & \textbf{1.83} & 3.61 & 156.5 & 6.4 & 706 & 6.74 \\
DANIEL & 2.37 & 4.99 & 23.8 & 41.6 & 2986 & 10.80 \\
\midrule
\multicolumn{7}{l}{\textit{Off-the-shelf systems}} \\
PERO-OCR & 13.50 & 26.19 & 30.9 & 32.4 & -- & -- \\
Tesseract v5 & 4.79 & 16.06 & 97.9 & 10.2 & -- & -- \\
TrOCR (zero-shot) & 15.46 & 60.97 & 69.5 & 14.4 & 1389 & -- \\
TrOCR (fine-tuned) & 3.62 & 8.95 & 284.3 & 3.5 & 1472 & 27.70 \\
Gemini & 7.23 & 19.11 & $\sim$1000 & $\sim$1 & -- & -- \\
\bottomrule
\end{tabular}
}
\end{table*}

\begin{table*}[ht!]
\small
\centering
\caption{Line-level recognition performance on Fraktur test set 
(Best in \textbf{bold}).}
\label{tab:line-fraktur}
\resizebox{\textwidth}{!}{
\begin{tabular}{@{}lcccccc@{}}
\toprule
\textbf{Model} & \textbf{CER }  & \textbf{WER }  & \textbf{Latency }  & \textbf{Throughput }  & \textbf{Inf Mem } & \textbf{Train Mem } \\
 & \textbf{(\%, $\downarrow$)}  & \textbf{(\%, $\downarrow$)}  & \textbf{(ms, $\downarrow$)} & \textbf{(img/s, $\uparrow$)} & \textbf{(MB)} & \textbf{(GB)} \\

\midrule
\multicolumn{7}{l}{\textit{Mamba-based (ours)}} \\
Mamba-CTC & 3.20 & 6.84 & 8.26 & 121.0 & 564 & 7.33 \\
Mamba-AR & 3.06 & 5.73 & 56.0 & 17.8 & 2915 & 8.01 \\
Mamba-NAR & 3.02 & 7.63 & 13.7 & 73.0 & 981 & 7.07 \\
\midrule
\multicolumn{7}{l}{\textit{Transformer baselines}} \\
VAN & \textbf{3.01} & 5.21 & \textbf{6.53} & \textbf{153.3} & 1473 & 16.27 \\
DAN & 3.03 & \textbf{4.88} & 166.0 & 6.0 & 703 & \textbf{6.10} \\
DANIEL & 6.18 & 14.11 & 23.7 & 41.9 & 3109 & 12.79 \\
\midrule
\multicolumn{7}{l}{\textit{Off-the-shelf systems}} \\
PERO-OCR & 15.73 & 32.44 & 33.2 & 30.1 & -- & -- \\
Tesseract v5 & 21.42 & 65.13 & 100.1 & 10.0 & -- & -- \\
TrOCR (zero-shot) & 35.98 & 98.40 & 72.0 & 13.9 & 1389 & -- \\
TrOCR (fine-tuned) & 4.71 & 10.79 & 414.5 & 2.4 & 1472 & 19.36 \\
Gemini & 10.50 & 29.59 & $\sim$1000 & $\sim$1 & -- & -- \\
\bottomrule
\end{tabular}
}
\end{table*}

\subsection{Paragraph-Level Recognition}
\label{subsec:para-results}

Table~\ref{tab:paragraph} presents results on severely degraded paragraph images. DAN achieves best accuracy (5.24\% CER), but VAN delivers the most practical profile: 6.42\% CER with 17.2 ms latency (23× faster than DAN). Mamba-AR maintains competitive accuracy (6.07\% CER) while outpacing DAN by 2.05× (195.6 ms vs 401.2 ms). DANIEL's efficiency advantage over DAN (112.1 ms vs 401.2 ms) demonstrates BPE tokenization benefits for reducing autoregressive sequence length. We report only Mamba-AR at paragraph level. Mamba-CTC and Mamba-NAR 
were designed for single-line recognition and lack multi-line layout 
modeling mechanisms, making them unsuitable for paragraph images 
containing stacked text lines.

\begin{table*}[t!h]
\small
\centering
\caption{Severely degraded paragraph-level recognition performance on BnL test set (976 paragraphs, 1--10 lines). Best results in \textbf{bold}.}
\label{tab:paragraph}
\resizebox{\textwidth}{!}{
\begin{tabular}{@{}lcccccc@{}}
\toprule
\textbf{Model} & \textbf{CER }  & \textbf{WER }  & \textbf{Latency }  & \textbf{Throughput }  & \textbf{Inf Mem } & \textbf{Train Mem } \\
 & \textbf{(\%, $\downarrow$)}  & \textbf{(\%, $\downarrow$)}  & \textbf{(ms, $\downarrow$)} & \textbf{(img/s, $\uparrow$)} & \textbf{(MB)} & \textbf{(GB)} \\
\midrule
\multicolumn{7}{l}{\textit{Neural models}} \\
Mamba-AR & 6.07 & 8.81 & 195.6 & 4.4 & 3021 & 14.56 \\
VAN & 6.42 & \textbf{7.64} & \textbf{17.2} & \textbf{58.1} & 620 & 8.80 \\
DAN & \textbf{5.24} & 9.14 & 401.2 & 2.5 & 948 & 19.17 \\
DANIEL & 6.18 & 9.53 & 112.1 & 8.9 & 3723 & 29.40 \\
\midrule
\multicolumn{7}{l}{\textit{Off-the-shelf systems}} \\
PERO-OCR & 12.58 & 19.34 & 35.8 & 27.9 & -- & -- \\
Tesseract v5 & 12.12 & 24.37 & 105.7 & 9.5 & -- & -- \\
Gemini & 6.06 & 9.65 & $\sim$1000 & $\sim$1 & -- & -- \\
\bottomrule
\end{tabular}
}
\end{table*}

\subsection{Memory Scaling Analysis}
\label{subsec:memory-scaling}
\begin{figure}[!t]
\centering
\includegraphics[width=1\textwidth]{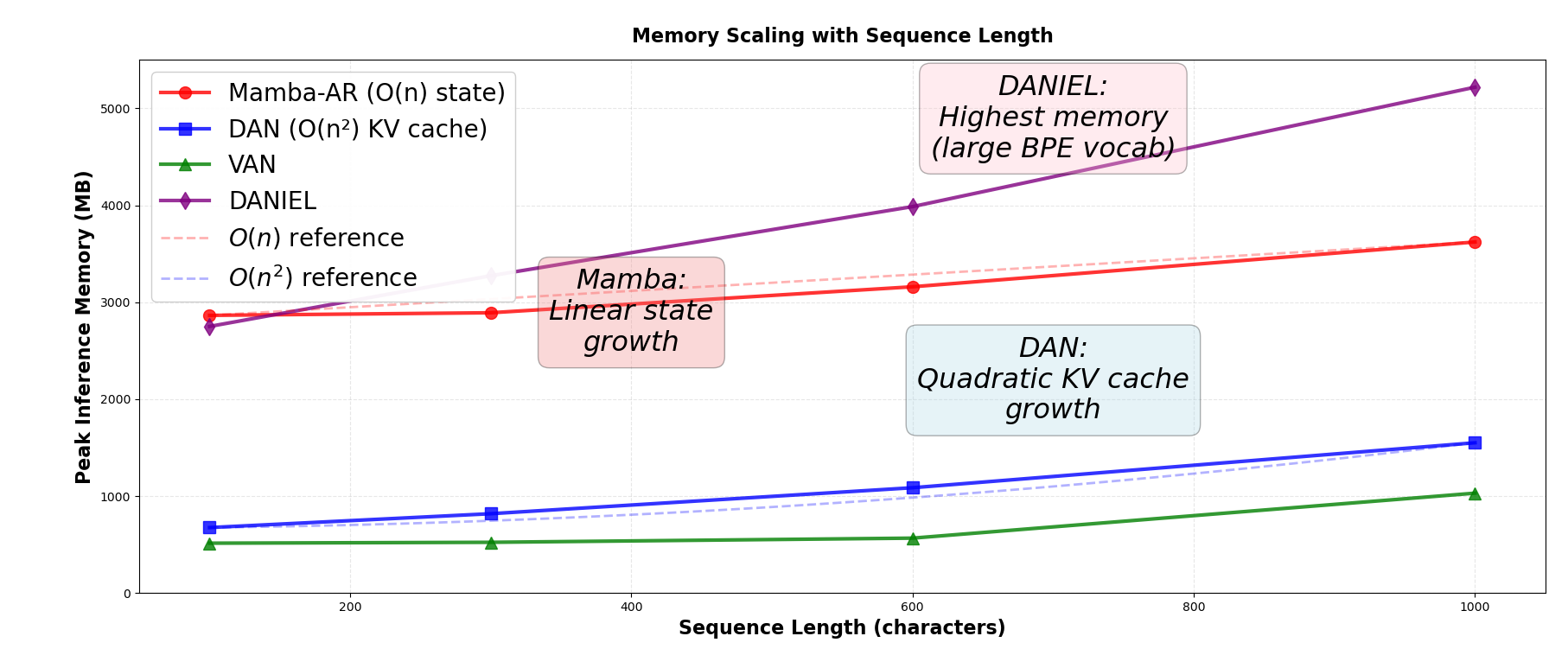}
\caption{Peak inference memory vs sequence length. Mamba-AR scales linearly ($O(n)$) while DAN quadratically ($O(n^2)$), confirming theoretical predictions.}
\label{fig:memory-scaling}
\end{figure}

We used synthetic data at controlled lengths (100/300/600/1000 chars) specifically to isolate the architectural scaling behavior from data-quality effects. On this synthetic data, all evaluated models achieve $\sim$1\% CER across all length bins, so the scaling story is cleanly about efficiency rather than accuracy degradation. 
Table~\ref{tab:memory-scaling} and Figure~\ref{fig:memory-scaling} examine peak inference memory as sequence length increases from 100 to 1000 characters. Mamba-AR exhibits near-linear growth (1.26× increase), while DAN scales quadratically (2.30× increase) due to expanding key-value caches. VAN shows favorable scaling until 600 characters, after which memory usage doubles (2.00×). DANIEL's growth (1.90×) reflects BPE compression reducing token count while still following Transformer scaling laws. These results empirically validate SSM's theoretical $O(n)$ complexity advantage: for 1000-character sequences.

 \vspace{-0.15cm}
 
\begin{table}[ht]
\centering
\caption{Peak inference memory scaling with sequence length. Mamba's higher baseline (2865 MB vs 676 MB) reflects larger parameter count (14.1M vs 7.1M), but exhibits linear $O(n)$ growth (1.26×) vs DAN's quadratic $O(n^2)$ (2.30×), enabling better scalability for long sequences.}
\label{tab:memory-scaling}
\footnotesize
\begin{tabular}{@{}lcccc@{\hspace{1.5em}}cccc@{}}
\toprule
\textbf{Model} & \multicolumn{4}{c}{\textbf{Peak Memory (MB)}} & \multicolumn{4}{c}{\textbf{Growth Factor (vs 100 chars)}} \\
\cmidrule(lr){2-5} \cmidrule(lr){6-9}
 & \textbf{100} & \textbf{300} & \textbf{600} & \textbf{1000} & \textbf{100} & \textbf{300} & \textbf{600} & \textbf{1000} \\
\midrule
Mamba-AR & 2,865 & 2,892 & 3,160 & 3,622 & 1.00× & 1.01× & 1.10× & 1.26× \\
DAN      & 676   & 819   & 1,087 & 1,551 & 1.00× & 1.21× & 1.61× & 2.30× \\
VAN      & 515   & 524   & 567   & 1,031 & 1.00× & 1.02× & 1.10× & 2.00× \\
DANIEL   & 2,751 & 3,274 & 3,987 & 5,219 & 1.00× & 1.19× & 1.45× & 1.90× \\
\bottomrule
\end{tabular}
\end{table}

\section{Conclusion}
\label{sec:conclusion}

This paper presents the first application of State-Space Models to historical document OCR, introducing three Mamba-based architectures with distinct decoding strategies (CTC, autoregressive, and non-autoregressive). Through systematic evaluation on the BnL newspaper corpus (8,328 lines, 976 annotated paragraphs, Antiqua and Fraktur scripts), we demonstrate that Mamba offers a compelling alternative to Transformers, particularly for paragraph-level recognition where computational efficiency is critical.

Mamba-AR achieves line-level accuracy comparable to DAN (1.83\% CER both) with 2.9× faster inference (53.3 ms vs 156.5 ms). At paragraph level, Mamba-AR delivers 2.05× speedup over DAN (195.6 ms vs 401.2 ms) while maintaining CER within 0.83\% (6.07\% vs 5.24\%). Memory scaling experiments empirically confirm Mamba's $O(n)$ complexity advantage over Transformers' $O(n^{2})$ at 1000-character sequences, memory grows 1.26× for Mamba versus 2.30× for DAN, validating theoretical predictions and enabling larger batch sizes for long-sequence processing.

Our findings reveal that while all neural models achieve competitive line-level accuracy ($\sim$2\% CER), computational efficiency differentiates them at paragraph level. VAN emerges as the optimal production baseline, delivering 58.1 img/s throughput with 6.42\% CER—23× faster than DAN with only 1.18\% CER penalty. Character-level tokenization suffices for historical text. BPE-based DANIEL's Fraktur collapse (6.18\% vs 2.37\% CER on Antiqua) demonstrates vocabulary mismatch risks. Foundation models (TrOCR, Gemini) underperform specialized systems despite massive pre-training, revealing persistent domain adaptation gaps for degraded historical documents.

We release our code, trained checkpoints, and BnL benchmark splits upon publication to establish a reproducible foundation for future research. As cultural heritage institutions digitize billion-page archives, this work guides practitioners toward architectures balancing quality with deployment feasibility, and invites further exploration of efficient sequence modeling for document understanding.
%
%
%
%
\bibliographystyle{splncs04}  
\bibliography{references}
\end{document}